# Autonomous Prompt Engineering in Large Language Models

**Daan Kepel**
Erasmus University
daankepel@gmail.com

**Konstantina Valogianni**
IE University
konstantina.valogianni@ie.edu

## Abstract

Prompt engineering is a crucial yet challenging task for optimizing the performance of large language models (LLMs) on customized tasks. This pioneering research introduces the Automatic Prompt Engineering Toolbox (APET), which enables GPT-4[1] to autonomously apply prompt engineering techniques. By leveraging sophisticated strategies such as Expert Prompting, Chain of Thought, and Tree of Thoughts, APET empowers GPT-4 to dynamically optimize prompts, resulting in substantial improvements in tasks like Word Sorting (4.4% increase) and Geometric Shapes (6.8% increase). Despite encountering challenges in complex tasks such as Checkmate in One (-14.8%), these findings demonstrate the transformative potential of APET in automating complex prompt optimization processes without the use of external data. Overall, this research represents a significant leap in AI development, presenting a robust framework for future innovations in autonomous AI systems and highlighting the ability of GPT-4 to bring prompt engineering theory to practice. It establishes a foundation for enhancing performance in complex task performance and broadening the practical applications of these techniques in real-world scenarios.[2]

## 1 Introduction

The landscape of artificial intelligence has undergone a remarkable transformation in recent years. In the past, leveraging AI for specific tasks required a dedicated team of data scientists to build and train specialized models. This process was not only resource-intensive but also limited in its accessibility to organizations with the requisite expertise and financial capacity. However, the advent of Large Language Models (LLMs) like GPT-4 has radically changed this scenario.

LLMs are advanced artificial intelligence systems designed to process, understand, and generate human language by learning from extensive datasets. Imagine a tool that can read and understand vast amounts of text—everything from books and articles to websites and social media posts. LLMs use this knowledge to perform a wide range of tasks involving language. They function as versatile tools capable of performing a broad range of linguistic tasks, from translation and content creation to answering complex questions, without requiring task-specific programming.

---

[1] Experiment funded by OpenAI Researcher Access Program.
[2] The datasets, the code and the model outputs are all available at https://github.com/daankepel/APET.

Through their ability to generalize across different domains, generalist foundation LLMs like GPT-4 represent a significant leap in AI. These models are outperforming specialized, state-of-the-art (SOTA) models right out of the box, without any need for task-specific training (OpenAI, 2023). This shift, together with the rise of ChatGPT as a product available to the public, which rose to 100 million users in just two months (Milmo, 2023), represents a significant democratization of AI, making powerful tools accessible to a wider audience.

The evolution of Large Language Models in recent years has been nothing short of revolutionary. This progress can be quantified in terms of the scale of model architecture and training data. The journey began with smaller-scale models like the original Transformer, introduced by Vaswani et al. (2017), which laid the groundwork for modern LLMs and enabled the creation of models like GPT-1, which were trained on datasets comprising of 117 million parameters (Radford et al., 2018), a figure that was groundbreaking at the time. These parameters refer to the internal settings of the model that are learned from the training data. These parameters help the model make predictions and generate text. The more parameters a model has, the more complex and nuanced its understanding and generation of text can be. The following years saw an exponential growth in the size and complexity of these models. BERT, short for Bidirectional Encoder Representations from Transformers, is a groundbreaking model in the field of natural language processing (NLP) introduced by Google in 2018. It revolutionized how machines understand human language by focusing on the context of words in a sentence, rather than just the words themselves (Devlin et al., 2018). Bidirectional training allows the model to understand the context of a word based on all of its surroundings (both left and right of the word), unlike previous models which processed text in one direction at a time. BERT quickly became a benchmark in NLP tasks, including being applied to Google Search (Google, 2019). After this, the development of more advanced LLMs accelerated. It was the release of GPT-3 in 2020, a model with 175 billion parameters, that set a new standard for LLMs. GPT-3's ability to understand context and generate coherent text on a wide range of topics was unprecedented (Brown et al., 2020).

The most recent milestone in this journey is OpenAI's GPT-4. This model, estimated at a staggering one trillion parameters, is five times the size of GPT-3 and approximately 3,000 times the size of BERT when it first came out. The sheer scale of GPT-4 represents a significant advancement in the field, with capabilities far surpassing its predecessors (OpenAI, 2023). The GPT-4 model can be described as a state-of-the-art foundation model, which is a term coined by researchers at Stanford University (Bommasani et al., 2021). Foundation Models are characterized by their scale, the extent of their training data, and their ability to be adapted to a wide range of tasks without task-specific training.

The most recent iterations of these models (i.e. GPT-4, and to some extent Google's PaLM) demonstrate emerging capabilities, including reasoning (the ability to make sense of complex information and come to logical conclusions), planning (the ability to sequence actions towards achieving a goal), decision-making (choosing between different options based on given criteria), in-context learning (adapting to new tasks based on the context provided without additional training), and responding in zero-shot scenarios (handling tasks they have never seen before without any prior examples).

These skills are attributed to their vast scale and the complexity of their training, despite the fact that the pretrained LLMs are not explicitly programmed to exhibit these attributes (Wei et al., 2022). It's important to note that all this is happening not because LLMs can actually think, but simply because they can generate text and have been trained on massive amounts of text. LLMs use deep neural networks, which are complex mathematical models inspired by the way the human brain works. These networks consist of layers of nodes (neurons) that process and transmit information. Through extensive training on large datasets, these networks learn to recognize patterns and relationships in the data, enabling LLMs to generate text that appears thoughtful and contextually appropriate. In these varied tasks, the performance of GPT-4 is remarkably close to that of a human expert, showcasing a near-human level of competence and adaptability (Bubeck et al., 2023). This near-human performance is an emergent property of the complex interactions within the neural network, the extensive training on diverse data, and the ability of the model to generalize from this data. Thus, while LLMs do not think in the human sense, their sophisticated architecture and training enable them to mimic many aspects of human-like reasoning and language use.

Despite these advancements, a critical challenge persists: the efficacy of LLMs is heavily dependent upon the quality of input prompts they receive (Wei et al., 2023). While a carefully crafted prompt can harness the full potential of these AI systems, an inadequately formulated prompt can yield results that fall short of their potential. This happens because LLMs generate responses based on the context provided by the prompts. A well-crafted prompt provides clear, specific, and relevant context, guiding the model to produce accurate and coherent responses. In contrast, a poorly designed prompt may lack clarity, specificity, or necessary context, leading the model to generate responses that are vague, irrelevant, or incorrect.

The reliance of the models on prompt design establishes a significant barrier, particularly for users who do not possess the expertise or experience in crafting effective prompts (Zamfirescu-Pereira et al., 2023). For instance, a prompt that ambiguously asks "Tell me about it" can lead to a variety of responses depending on what "it" refers to, whereas a more specific prompt like "Explain the process of photosynthesis in plants" is likely to yield a focused and accurate explanation. Additionally, the use of structured prompts that guide the model through a step-by-step process or include specific instructions can significantly enhance the quality of the output.

Consequently, the democratization of AI, with all its potential, faces limitations in its depth of accessibility and utility. Without the ability to formulate effective prompts, many users may find it challenging to fully leverage the capabilities of LLMs. This underscores the importance of developing tools and methodologies to assist users in creating high-quality prompts, thereby making powerful AI technologies more accessible and effective for a broader audience.

Recent literature has explored various methods to improve the performance of LLMs through prompt optimization. Studies have introduced techniques such as Chain of Thought (CoT) prompting, Tree of Thoughts (ToT) frameworks, and self-consistency methods to enhance the reasoning and decision-making abilities of LLMs (Wei et al., 2022; Yao et al., 2023). Additionally, research has focused on methods like "Ask Me Anything" (AMA) and universal prompt retrieval systems (UPRISE) to improve zero-shot performance and reduce hallucinations (Arora et al., 2022; Cheng et al., 2023). These advancements have significantly improved the accuracy and reliability of LLMs, yet they still largely depend on human-crafted prompts and external interventions. To address these challenges and enhance the capabilities of the LLM, this research aims to explore the autonomous capabilities of GPT-4, focusing on its potential to self-optimize prompts. Self-optimization refers to the ability of a system, in this case, GPT-4, to autonomously refine and improve the prompts it receives to generate more accurate and relevant responses. This involves the model analysing the initial prompt, identifying potential improvements, and adjusting the prompt to better suit the task at hand. Moreover, we explore the increasing ability of generalist foundation models to walk the fine line between specialized expertise and broad applicability of these AI models.

The theoretical contributions of this research are significant, enhancing the body of knowledge on the autonomous capabilities of Large Language Models like GPT-4. This study aims to advance understanding of how such models can independently optimize prompts, challenging the current reliance on human intervention for improving AI performance. It suggests a move towards LLMs that can self-improve, broadening the research into AI's potential for self-directed learning and adaptation.

From a practical standpoint, the implications of this research extend into the wider adoption and application of AI technologies across diverse sectors. By demonstrating GPT-4's ability to autonomously optimize prompts, this study highlights the potential for LLMs to lower the barriers to effective AI use, making sophisticated AI tools more accessible to non-experts. This democratization of AI could revolutionize how businesses, educational institutions, and individuals approach problem-solving, creativity, and decision-making, fostering innovation and efficiency. Furthermore, the insights derived from this research could inform the development of more intuitive and self-sufficient AI systems, paving the way for broader societal adoption of AI technologies. In doing so, this study not only contributes to the academic discourse but also offers practical strategies for harnessing the full potential of LLMs in real-world applications.

## 2    Literature Review

This literature review critically examines the evolution and current state of Large Language Models (LLMs) and their role in the democratization of artificial intelligence. It will trace the development of these models from their early iterations to the advanced, multifaceted GPT-4, highlighting key technological advancements and their implications. A particular focus will be on the challenges and strategies related to prompt design, its impact on the effectiveness of LLMs and the consequences that this has for every day users. The review will also explore recent studies on prompt optimization within these models. By synthesizing existing research, this review aims to contextualize the challenges and potentials of LLMs, setting the stage for the research questions addressed in this thesis.

## 2.1     The Evolution and Inner Workings of Large Language Models (LLMs)

The evolution of LLMs is an important aspect of the modern artificial intelligence landscape, characterized by a series of revolutionary advancements in model architecture, training techniques, and an increasingly sophisticated understanding of language.

At the heart of this evolution is the Transformer model, introduced by Vaswani et al. in 2017. This model marked a significant departure from previous approaches in natural language processing (NLP) through its unique use of the 'attention mechanism', which was first introduced by Bahdanau et al. (2015). Unlike earlier models that processed input sequences in a linear or sequential manner, the Transformer could focus on different parts of the input sequence, determining which parts were most relevant for a given task. This attention mechanism is akin to how a spotlight highlights specific actors on a stage, allowing the audience to focus on key performances while maintaining awareness of the entire scene.

Following the Transformer, OpenAI developed the Generative Pre-training Transformer (GPT) series, starting with GPT-1 (Radford et al., 2018). This model leveraged the Transformer architecture to generate coherent text, demonstrating the potential of scaling up models for improved performance. These models operate using tokens, which are essentially pieces of text converted into a format understandable by the model. The tokens are processed through layers of neural networks – a complex arrangement of nodes and connections inspired by the human brain's architecture. Each layer of the network captures different aspects of language, from basic syntax to complex semantic relationships.

BERT, introduced by Google, added another dimension to this landscape with its bidirectional training approach (Devlin et al., 2019). Unlike the unidirectional approach of GPT models, where the context is understood based on preceding text, BERT analyzes text in both directions – forwards and backwards. This bidirectionality allows for a more nuanced understanding of context, as the meaning of a word can be influenced by words that come both before and after it.

The release of GPT-3 by OpenAI took these advancements further, scaling up the model to unprecedented levels (Brown et al., 2020). With an increased number of parameters and more extensive training data, GPT-3 was capable of generating even more nuanced and contextually aware text. Its successor, GPT-4, continued this trend, pushing the boundaries of model size and complexity, resulting in enhanced linguistic proficiency and a broader range of capabilities (OpenAI, 2023).

LLMs operate on complex concepts such as word embeddings, transformer architecture, and self-attention mechanisms. Word embeddings translate words into high-dimensional vectors, capturing semantic relationships. Transformers process these embeddings, using self-attention to weigh the importance of different words in a sentence for understanding context. The attention mechanism involves assigning weights to different parts of the input data, which are calculated using SoftMax functions – a mathematical approach that helps the model decide which parts of the input to focus on. The training of these models involves adjusting model parameters to minimize the difference between predicted and actual word sequences, refining the model's ability to generate coherent and contextually relevant text. (Lee & Trott, 2023).

The combination of these mathematical concepts enables LLMs like GPT-4 to perform a wide range of tasks, from generating human-like text to understanding and translating languages, answering questions, and even creating content like poetry or code (Naveed et al., 2023). The models do so by effectively learning patterns and relationships in the data they are trained on, mirroring, to some extent, the way humans learn and understand language.

## 2.2   Performance and Evaluation of Foundation LLMs

Foundation models are transformative AI systems trained on extensive datasets to grasp a broad spectrum of knowledge, enabling them to be adapted for diverse tasks without domain-specific tuning (Bommasani et al., 2022). These models, including GPT-4, BERT, and others, through self-supervised learning from extensive data, demonstrate adaptability to numerous tasks (OpenAI, 2023). This chapter delves into the comparative analysis of the advancements in foundation models across various domains. By examining their capabilities, limitations, and potential for innovation, we aim to map the current AI landscape and lay the groundwork for this research's further development.

As the current landscape of AI development has been evolving with incredible speed, we will compare the latest, state-of-the-art models with each other, evaluating the LLMs on specific tasks. Two broader categories of tasks have been defined to evaluate the performance of Language Models (Naveed et al., 2023):

1. **Natural Language Understanding (NLU)**: This task evaluates the language comprehension abilities of Language Models. It covers a range of tasks such as sentiment analysis, text classification, natural language inference, question answering, commonsense reasoning, and reading comprehension, among others.
2. **Natural Language Generation (NLG)**: This task covers the language production proficiency of Large Language Models based on the given input context. It involves tasks like summarization, sentence completion, machine translation, and dialogue generation, among others.

Next to these LLM capabilities, it is evident that the scale of the latest generation LLMs, notably, GPT-4 (OpenAI, 2023), PaLM (Anil et al., 2023), and LLaMa (Touvron et al., 2023), has uncovered emerging capabilities. These are tasks that a smaller size LLM is not able to perform, but only emerges once the size of the model (e.g., training compute, model parameters, etc.) becomes large enough (Wei et al., 2022). These emergent abilities include performing arithmetic, playing chess, summarizing passages, and more, which LLMs learn simply by observing natural language. Moreover, it becomes increasingly apparent that the increasing scale of these models not only reveals new capabilities but also significantly enhances their proficiency in the described tasks (Srivastava et al., 2023). This enhancement in performance underscores the critical role that model scale plays in advancing what is achievable with AI, making these state-of-the-art LLMs more versatile across a broader spectrum of complex tasks.

Measuring the performance of these Large Language Models is currently done by a vast list of benchmark datasets. The benchmark datasets measure the natural language processing, general knowledge, reasoning, and problem-solving capabilities of these models. Benchmarks such as GLUE (Wang et al., 2019) and SuperGLUE (Wang et al., 2020) challenge models on a range of natural language understanding tasks, including sentiment analysis and paraphrase detection, while specialized datasets like ARC (Moskvichev et al., 2023) and MMLU (Hendrycks et al., 2020) go deeper into models' reasoning capabilities and general knowledge across various disciplines. More advanced tasks presented by benchmarks like AGIEval (Zhong et al., 2023) and BIG-Bench (Srivastava et al., 2022) test the limits of LLMs' problem-solving abilities, including their capacity for multi-step reasoning and understanding complex, real-world scenarios. As LLMs continue to evolve, these benchmarks serve as critical tools for assessing their progress, highlighting both their strengths and limitations. However, measuring the performance of these models is getting increasingly difficult, as traditional benchmarks may no longer fully capture the breadth and depth of capabilities exhibited by state-of-the-art LLMs like GPT-4 (Bubeck et al., 2023).

The challenge lies in the models' ability to generalize across a vast array of tasks, some of which have not been explicitly encountered during training. This generality suggests a form of intelligence that transcends simple pattern recognition or memorization, venturing into areas of creativity, problem-solving, and even intuition. Given this context, the evaluation of such models demands innovative approaches that go beyond conventional metrics. Collins et al. (2022) and Bubeck et al. (2023) propose methods which focus more on linguistic reasoning, measuring the ability to perform human-like tasks (e.g. the ability to plan, the ability to explain why something is happening). Next to these methods, other benchmarking datasets have also been developed which focus more on the emerging capabilities of these LLMs. For instance, The Game of 24 (Yao et al., 2023), challenges LLMs to creatively form arithmetic expressions to reach the number 24 from four given numbers, testing their numerical reasoning in a unique manner. Similarly, the BIG-Bench Hard (BBH) tasks introduced by Suzgun et al. (2023), including Geometric Shapes, Multi-Step Arithmetic Two, and Word Sorting, along with a reasoning task, Checkmate-in-One. Python Programming Puzzles (P3) by Schuster et al. (2021) presents a series of intricate programming challenges, assessing LLMs' coding prowess across various difficulty levels. Additionally, the Multilingual Grade School Math (MGSM) dataset by Shi et al. (2023) expands the scope of evaluation to include linguistic diversity, translating arithmetic problems into ten different languages, thereby testing not just mathematical logic but also cross-linguistic understanding. While more resource intensive, these methods cover the ability of these models to perform tasks beyond Natural Language Understanding and Generation.

## 2.3 The Role of Prompt Design

As the performance of these models increases as the size keeps growing, the efficacy of LLMs is not solely a manner of their architectural design or the size of their training data. An equally critical factor is the manner in which these models are interacted with, particularly through the formulation of prompts. This chapter explores the important role of prompting techniques in unlocking the full potential of LLMs, highlighting how it has become a cornerstone in the practical application of these advanced AI tools.

The significance of prompt engineering stems from its direct impact on the quality, relevance, and accuracy of the responses generated by LLMs. A well-optimized prompt can lead to outputs that closely align with user expectations, effectively leveraging the model's capabilities. In contrast, poorly crafted prompts may yield irrelevant or inaccurate responses (Zhou et al., 2022).

Prompt engineering refers to the process of crafting inputs for LLMs in a way that effectively guides the models to generate the desired outputs (Chen et al., 2023). Given the generalist nature of LLMs, which are not fine-tuned for specific tasks, the use of prompt engineering emerges as a crucial skill for users and developers alike. It enables a more intuitive interaction with AI, transforming these models from mere repositories of information into dynamic tools capable of engaging in creative problem-solving, generating insightful analyses, and even participating in complex decision-making processes (Bubeck et al., 2023). Moreover, using the correct prompting techniques even enables generalist LLMs like GPT-4 to outperform fine-tuned models, specifically trained for a task (Nori et al., 2023). This means that, using the correct formulation of a question, GPT-4 can outperform fine-tuned models, further contributing to the democratization of AI.

Moreover, prompt engineering is not a static field; it is continuously evolving in response to advances in model architectures, changes in user needs, and the development of new application areas. Researchers and practitioners are exploring various strategies to refine the process, including the use of prompt engineering, few-shot learning (examples of correct answers), and the incorporation of meta-information into prompts, which refers to additional context or data about the primary information within a prompt that can help guide the model's response. These efforts are aimed at developing more systematic and efficient ways to interact with LLMs, making AI more accessible and effective for a broader audience.

To further investigate the different prompting techniques, we first make a distinction between zero-shot prompting and few-shot prompting. Zero-shot and few-shot prompting are techniques used in machine learning, particularly with language models, to handle tasks without or with minimal task-specific training data. Zero-shot prompting requires a model to perform a task without any prior examples, relying on its pre-existing knowledge and the instructions provided within the prompt. It assesses the model's ability to generalize from its training to new tasks without explicit examples (Wang et al., 2019). This capability is essential for tasks where labeled data is scarce or not available. In contrast, few-shot prompting involves providing the model with a small number of examples (the "shots") within the prompt that illustrate what is expected. These examples serve as a direct guide, helping the model understand the context and the specific task requirements. Few-shot prompting effectively leverages the model's learned patterns from training and applies them to the task at hand with the help of these examples, enhancing its ability to generate more accurate and relevant responses based on the limited examples provided (Xia et al., 2020). Understanding the distinction between these two approaches helps us tailor the model to specific applications, where data might not be freely available but reasoning capabilities are still needed.

Prompt optimization techniques for enhancing the performance of models vary widely in complexity. Simple strategies include the use of delimiters to separate different sections of the input clearly (OpenAI, n.d.), which can help in structuring the information more effectively for the model. Even seemingly straightforward interventions, such as prefacing prompts with phrases like "Let's think step by step," have been shown to significantly boost the model's performance in a zero-shot environment (Kojima et al., 2023). On the more complex end of the spectrum, there are multi-step approaches that necessitate multiple interactions with the model to refine the response.

One method that further underscores this advancement is the Chain of Thought (CoT) approach. Chain of Thought prompting has emerged as a compelling method for enhancing the complex reasoning capabilities of LLMs. This technique involves providing models with prompts that include a series of intermediate reasoning steps, which guide the model towards generating the final answer. Studies have shown that when models are prompted within a few-shot environment demonstrating this chain of thought, their performance improves significantly on various arithmetic, commonsense, and symbolic reasoning tasks. For instance, the use of CoT prompting with just eight examples has enabled a PaLM 540B model to achieve state-of-the-art accuracy on the GSM8K benchmark, a collection of math word problems, outperforming even fine-tuned GPT-3 models equipped with a verifier (Wei et al., 2022). Whether it's solving mathematical puzzles or making logical deductions, the CoT approach not only elevates the accuracy of the outcomes but also renders the model's thought process transparent and understandable to users.

The "Tree of Thoughts" (ToT) framework introduced by Yao et al. (2023) expands on the CoT method by enabling LLMs to explore multiple reasoning paths and evaluate different solutions to solve complex problems. ToT allows for strategic decision-making, looking ahead, and backtracking when necessary, significantly enhancing LLMs' problem-solving abilities across tasks like the Game of 24, Creative Writing, and Mini Crosswords. For example, ToT achieved a 74% success rate in the Game of 24, a substantial improvement over CoT's 4% success rate with GPT-4. This framework represents a novel approach to leveraging LLMs for extended problem solving and reasoning.

A method that is more focused on the reliability of the output, is "Self-Consistency". This method generates multiple reasoning paths and selects the most consistent answer across them, leveraging the intuition that correct reasoning often follows multiple paths to the same conclusion. Self-consistency significantly boosts performance across various arithmetic and commonsense reasoning benchmarks. This approach simplifies existing methods by working off-the-shelf with pre-trained models, requiring no additional training or human annotations, acting as a self-ensemble to enhance reasoning accuracy (Wang et al., 2022).

Recent research in prompt engineering has introduced sophisticated techniques aiming for more precise and contextually relevant outputs. A prominent innovation is the "Expert Prompting" method developed by Xu et al. (2023), which improves responses by first creating an "expert identity" aligned with the query's context, and then integrating this identity into the prompt. This method exists in two forms: a static version, which uses a consistent expert profile, and a dynamic version, creating a unique expert identity for each query to produce adaptive and finely tuned responses. Additionally, Du et al. (2023) managed to increase LLM performance through "Multi-persona Prompting," also referred to as solo-performance prompting (SPP). This approach directs the LLM to construct various "personas" tailored to a specific task or question. These personas participate in a simulated group discussion, offering solutions, critiquing each other, and refining their suggestions collaboratively. The final step synthesizes these interactions into a unified, comprehensive answer.

In conclusion, this chapter has highlighted the significance of prompt design in enhancing the performance of LLMs. The discussion underscored that beyond the model's architecture and training data size, the art of crafting prompts plays an important role in leveraging the full capabilities of LLMs. Through methodologies like Chain of Thought, Tree of Thoughts, and Expert Prompting, we have seen the potential for nuanced interaction between humans and AI to produce more accurate, relevant, and sophisticated outputs.

As we proceed to the next chapter, the focus shifts to the field of automated prompt optimization. This area represents an important research direction, aiming to reduce the reliance on manual prompt engineering by developing algorithms capable of refining and generating prompts autonomously. This advancement holds the promise of making LLMs more accessible and effective, by systematically improving how these models interpret and respond to user queries.

## 2.4    Emerging Trends in Prompt Optimization

As we have observed, the capacity of LLMs to interpret and respond to human queries with high degrees of accuracy and relevance is significantly influenced by the quality of the prompts they are given. This has led to an increased focus on developing methods that not only enhance the effectiveness of prompts but also enable models to autonomously refine their responses. These methods showcase a range of approaches, from enhancing model responsiveness with diverse prompts to enabling models to critique and improve their reasoning through debate, thereby marking significant progress in making LLMs more adaptable and reliable.

Arora et al. (2022) introduced a method called "Ask me Anything" (AMA) to improve the performance of language models on a variety of tasks without additional training. AMA involves using multiple, imperfect prompts and aggregating their outputs to produce a final prediction. The approach is based on the observation that question-answering (QA) prompts, which encourage open-ended responses, tend to be more effective than those that limit responses to specific formats. The authors develop a scalable method to transform task inputs into effective QA formats using the language model itself and then aggregate these using a process called weak supervision. This process combines noisy predictions to produce a final output. AMA was evaluated across different model families and sizes, demonstrating significant performance improvements over baseline models. The method enabled a smaller, open-source model to match or exceed the performance of larger, few-shot models on several benchmarks.

Cheng et al. (2023) present an approach to enhance the zero-shot performance of LLMs through a universal prompt retrieval system. This system, named UPRISE, employs a lightweight and versatile retriever that automatically selects the most effective prompts for any given task input in a zero-shot environment. The innovation of UPRISE lies in its ability to generalize across different tasks and models without the need for task-specific fine-tuning or manual prompt engineering. The retriever is trained on a diverse range of tasks but is capable of working on unseen tasks and with various LLMs, demonstrating its universal applicability. The methodology involves tuning the retriever on a smaller model (GPT-Neo-2.7B) and evaluating its effectiveness on larger models such as BLOOM-7.1B, OPT-66B, and GPT3-175B. Remarkably, UPRISE also shows potential in mitigating the hallucination issue prevalent in models like ChatGPT, thereby enhancing the factual accuracy of their outputs. This approach significantly improves upon baseline zero-shot performance across multiple LLMs and tasks, underscoring its potential to make LLMs more versatile and effective in real-world applications without extensive retraining.

Du et al. (2023) explore an approach to enhance the reasoning and factuality LLMs. The authors propose a method where multiple instances of LLMs engage in debates over their generated responses to a given query. This debate process involves iterative rounds where each LLM critiques and revises its responses based on the feedback from other models, aiming for a consensus. This method significantly improves the LLMs' ability to reason and generate factually accurate content across various tasks, including arithmetic, strategic reasoning (e.g., chess move prediction), and factual information extraction (e.g., generating biographies). The study demonstrates that this multiagent debate approach not only reduces the occurrence of false facts but also enhances the models' mathematical and strategic reasoning capabilities. The approach requires only black-box access to the LLMs, and shows that a "society of minds" can effectively advance LLMs' performance without the need for additional training or fine-tuning on specific tasks.

Another method, called AutoHint, combines the strengths of zero-shot and few-shot learning. It optimizes prompts by generating and incorporating hints from input-output demonstrations, significantly improving task accuracy. The method starts with an initial prompt, identifies incorrect predictions, and uses these to generate hints that refine the prompt. Evaluated on the BIG-Bench Instruction Induction dataset, AutoHint showed notable accuracy improvements across various tasks, demonstrating its effectiveness in prompt optimization (Sun et al., 2023).

PromptAgent is an optimization method that autonomously develops prompts of a level comparable to those designed by experts. PromptAgent employs a Monte Carlo tree search algorithm to explore and optimize the prompt space efficiently, leveraging error feedbacks to iteratively refine prompts towards expert-level quality through a process of selection, expansion, simulation, and back-propagation. This approach enables PromptAgent to generate highly effective, domain-specific prompts, demonstrating superior performance over strong Chain-Of-Thought methods across diverse tasks, including BIG-Bench Hard and various NLP challenges.

Another method, exploring the possibility of chaining LLM prompts to enhance the effectiveness of AI in complex tasks, allows the output of one LLM operation to serve as the input for the next, improving task outcomes and user experience in terms of transparency, controllability, and collaboration. Through a 20-person study, the authors demonstrate how chaining can lead to better quality results and offer users new ways to interact with LLMs, such as through calibration of model expectations and debugging of model outputs (Wu et al., 2022).

Cumulative Reasoning is a method that improves LLMs' ability to tackle complex problems through cumulative and iterative processing, emulating human thought processes. By breaking down tasks into smaller, manageable components, Cumulative Reasoning significantly enhances problem-solving effectiveness. The method employs three types of LLMs—proposer, verifier, and reporter—to progressively refine solutions, demonstrating superior performance on logical inference tasks and establishing new benchmarks on datasets like FOLIO wiki and MATH. This approach addresses the limitations of LLMs in handling complex tasks by facilitating a more structured and effective problem-solving process.

Automatic Prompt Engineering is a methodology that leverages LLMs for generating and selecting effective prompts automatically. This approach significantly enhances the performance of LLMs across a variety of NLP tasks by optimizing instructions to achieve better or comparable results to those generated by human experts. APE demonstrates its effectiveness by outperforming the baseline LLM performance and matching or exceeding human-level prompt engineering in most tasks, highlighting the potential of LLMs in reducing the manual effort involved in prompt design (Zhou et al., 2023).

In a groundbreaking study, Medprompt, employing dynamic few-shot selection, self-generated chain of thought, and choice shuffle ensembling, was introduced to significantly enhance GPT-4's performance on medical benchmarks. Without specialized training, these techniques combined to surpass existing benchmarks, demonstrating a remarkable 27% reduction in error rate on the MedQA dataset (Nori et al., 2023). This approach not only set new standards for accuracy but also showcased its broad applicability beyond medical domains, signaling a major advancement in the use of generalist models for specialized tasks.

Combining previous methods, Suzgun and Kalai (2024) introduce a technique called "meta-prompting". Meta-prompting transforms a singular LLM into a conductor, who can orchestrate multiple independent LLMs to collaboratively address complex tasks. This process involves the LLM breaking down tasks into smaller, manageable subtasks, which are then delegated to specialized "expert" instances of the same LLM, each provided with tailored instructions for execution. The central LLM, acting as the conductor, ensures seamless integration and communication among these expert models, applying critical thinking and robust verification processes to refine and authenticate the final outcome. Remarkably, meta-prompting outperformed standard prompting methods by significant margins, demonstrating an average improvement of 17.1% over standard prompting, 17.3% over dynamic expert prompting, and 15.2% over multi-persona prompting.

Ye et al. (2024) propose a novel approach called "Prompt Engineering a Prompt Engineer", which involves creating a meta-learning framework where the LLM is trained to optimize its own prompts. This technique allows the model to generate and refine its prompts iteratively, enhancing its performance across various tasks. It leverages historical task data to inform the prompt optimization process, ensuring that the generated prompts are contextually relevant and tailored to the task requirements. The method was tested on diverse benchmarks, including natural language understanding and generation tasks, showing substantial gains in performance and adaptability.

Pryzant et al. (2023) introduce "Gradient Descent for Prompt Optimization" (GDPO), a technique that applies gradient descent algorithms to refine prompts using existing task data. This approach treats prompt tokens as parameters that can be optimized to minimize the loss on specific tasks. By iteratively adjusting these tokens based on past performance data, GDPO fine-tunes the prompts to enhance the model's accuracy and efficiency. The authors evaluated GDPO across multiple benchmarks, including sentiment analysis and question answering, demonstrating notable improvements in task performance. This method highlights the potential of using traditional optimization techniques in the context of prompt engineering to achieve better results with LLMs.

The advancements in prompt optimization have laid a foundational framework, enabling LLMs to generate more accurate, contextually relevant responses based on enhanced prompts.

Previous research has laid a robust foundation for this study, illustrating the diverse methodologies aimed at optimizing prompt effectiveness and enabling LLMs to refine their outputs. The innovative approaches introduced by Arora et al. (2022), Cheng et al. (2023), Du et al. (2023), and others have significantly advanced our understanding of how to harness the full potential of LLMs. Notably, the research conducted by Nori et al. (2023) and the meta-prompting technique introduced by Suzgun and Kalai (2024) represent cutting-edge advancements in the field. They have shown that even without specialized training, LLMs can achieve and surpass benchmarks in highly specialized domains such as healthcare, through innovative prompt optimization and collaborative model interaction strategies. These studies have demonstrated that through methods such as AMA, UPRISE, multiagent debates, AutoHint, RAIN, SIRLC and Cumulative Reasoning, it is possible to markedly improve the responsiveness, reasoning, factuality and overall performance of LLMs across a wide range of tasks and domains, building a solid foundation for this study.

## 2.5 Research Contribution

The development and refinement of LLMs like GPT-4 represent a monumental shift in the capabilities of artificial intelligence, moving towards systems that can understand and generate human-like text across a broad spectrum of tasks without the need for domain-specific tuning. This evolution, as detailed in foundational works by Bommasani et al. (2021) and the groundbreaking capabilities showcased by Bubeck et al. (2023) and Nori et al. (2023), underscores the transformative potential of foundation models. These models have not only demonstrated remarkable linguistic proficiency but have also shown an ability to engage in complex reasoning, creative generation, and problem-solving tasks, setting a new benchmark for what is achievable with AI.

In parallel, the literature has increasingly recognized the critical role of prompt design in leveraging the full capabilities of LLMs. Innovative techniques such as "Expert Prompting" and "Multi-persona Prompting" (Xu et al., 2023; Du et al., 2023) have highlighted the potential for significantly enhancing model outputs through refined and optimized prompts. These studies illustrate the importance of the interaction between the user and the model, showcasing how carefully crafted prompts can lead to improved accuracy and relevance in the model's responses.

Despite these advancements, the literature reveals a significant gap in the autonomous operation of LLMs, particularly concerning self-optimization of prompts. Current research has largely focused on external methods for prompt optimization, overlooking the potential for models to internally refine prompts based on their understanding and processing capabilities. This gap highlights a crucial area for exploration, as autonomous prompt optimization could further democratize AI by making sophisticated models more accessible and intuitive for users. This brings forward the first three hypotheses:

**Hypothesis 1 (H1)**: *GPT-4 improves output quality significantly with self-optimized prompts versus unoptimized prompts.*

**Hypothesis 2 (H2)**: *GPT-4's self-produced prompt optimization yields performance on par with that of specialized external prompt optimization methods.*

**Hypothesis 3 (H3)**: *The benefits of self-produced prompt optimization by GPT-4 are consistent across all prompt types.*

This research experiment aims to directly address these gaps by exploring the feasibility of GPT-4 autonomously optimizing prompts. By investigating the model's capacity for internal prompt refinement, this study seeks to uncover the mechanisms through which GPT-4 can enhance its interactions with users autonomously. The feasibility of developing a self-contained system for prompt optimization within GPT-4 is supported by the model's existing capabilities for complex task performance, reasoning and natural language understanding. The literature provides a foundation upon which this experiment builds, arguing that the advanced cognitive and processing abilities of GPT-4 make it a suitable candidate for such autonomous operations.

Furthermore, this experiment extends beyond the current literature by combining the capabilities of prompt optimization into a single, coherent system within a foundation model. By investigating the model's performance across different types of prompts and tasks, this research broadens the understanding of LLM applicability and utility, demonstrating the potential for these models to serve a wider array of user needs and contexts autonomously. This brings forward hypothesis 5, which represents the overall goal of our research:

**Hypothesis 4 (H4)**: *GPT-4 can operate as a self-contained system, capable of optimizing prompts and generating answers autonomously.*

In summary, this experiment contributes to the existing body of literature by bridging identified gaps and pushing the boundaries of what is currently understood about the autonomous capabilities of Large Language Models. Through this research, we aim to pave the way for the next generation of AI systems, characterized by their ability to self-optimize and continuously improve, thereby making powerful AI tools more intuitive and accessible to all users.

## 3   Methodology

This chapter outlines the methodological framework for evaluating the autonomous capabilities of GPT-4. Grounded in the extensive literature review, this experiment aims to empirically test the model's ability to self-optimize and self-generate responses to various prompts, as visualized in the provided experiment design diagram (Figure 1). As we are focused on improving the general applicability of these models, we will focus on prompts in a zero-shot environment, where additional, labeled data for the prompt might not be freely available.

### 3.1   Research Design

The experiment follows a systematic approach, beginning with prompts from benchmarking datasets, which will be subject to our prompt optimization process, later dubbed the Autonomous Prompt Engineering Toolbox.

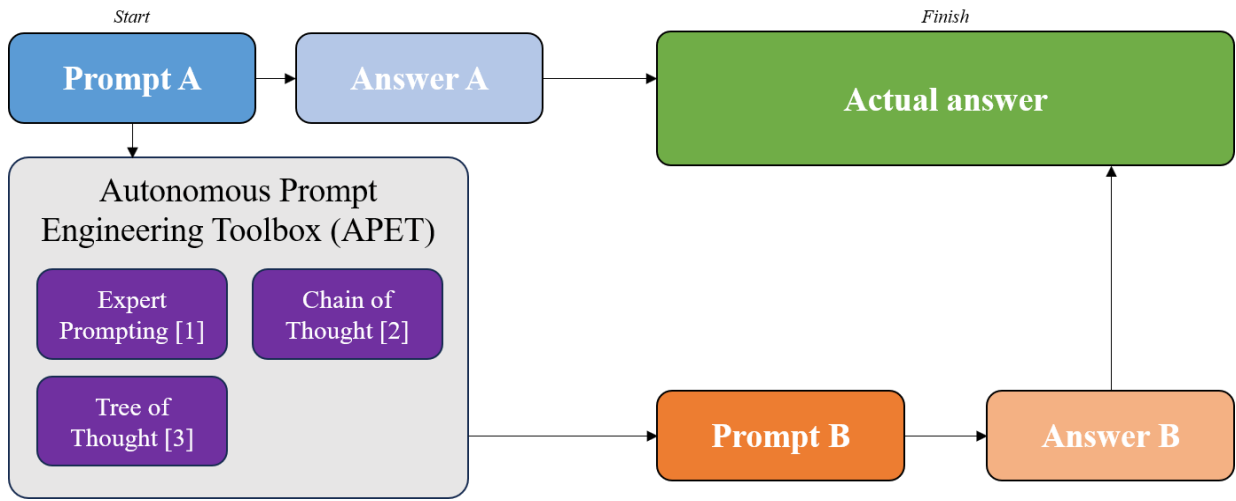

*Figure 1: Research Design*

Prompts will be sourced from benchmark datasets designed to test the model across a range of domains. These initial prompts will serve as the baseline for assessing the optimization capabilities of GPT-4. The optimization model will then be executed through OpenAI API calls, transforming the original prompt into an optimized one. GPT-4 will then generate two sets of responses: Answer A from the original prompt and Answer B from the optimized prompt. Both sets will be produced through separate OpenAI API calls to ensure an unbiased generation process. The answers are then compared to the answer provided by the benchmark dataset to assess the performance of both the unoptimized and optimized prompts.

## 3.2 Benchmark Datasets

To rigorously assess the effectiveness of the prompt optimization framework described in earlier sections, we utilize a variety of benchmark datasets. These datasets are designed to test the model's ability to handle complex logical, mathematical, and language tasks under different conditions.

The benchmark datasets chosen for this study encompass a range of tasks that challenge the model's reasoning, comprehension, and problem-solving capabilities in diverse contexts. Each dataset has been selected for its relevance to the specific aspects of LLM performance we aim to evaluate, as well as for their recognized rigor and utility in the AI research community. The following datasets form the core of our evaluation framework:

- **Checkmate in One (BIG-Bench authors, 2023)**: This dataset tests the model's ability to solve complex spatial and logical problems within the domain of chess. The task requires the LLM to determine a winning move that results in checkmate in one turn, challenging the model's strategic thinking and visualization skills.
- **Word-sorting (Suzgun et al., 2023)**: This dataset assesses the LLM's capability to sort words alphabetically. This task tests the model's understanding of alphabetical order and its ability to organize linguistic information systematically, providing insights into its processing efficiency and accuracy.

- **Game of 24 (Yao et al., 2023)**: This dataset involves arithmetic and number theory challenges where the model must manipulate four numbers using basic arithmetic operations to achieve a total of 24. This task evaluates the model's numerical reasoning skills and its ability to engage in complex problem-solving under constraints.
- **Geometric Shapes (Suzgun et al., 2023)**: This dataset evaluates a model's ability to identify various shapes from their SVG (Scalable Vector Graphics) path descriptions. Tasked with interpreting and classifying complex geometric data encoded in a text-based format, the model needs to leverage its understanding of SVG syntax and geometric principles. This challenge tests the model's capabilities in both visual interpretation and textual data processing, providing insight into its ability to integrate graphical information within a linguistic framework.

The actual datasets are sourced from Suzgun, M., & Kalai, A. T. (2024), who have provided a GitHub repository with the datasets used in their study.

## 3.3  Conceptual Framework of the Prompt Engineering Toolbox

The toolbox is crafted to enable LLMs to autonomously refine and improve upon given input prompts. This framework is designed to align input prompts with the model's processing strengths, optimizing[3] the prompts for accuracy, relevance, and depth.

### 3.3.1  The Autonomous Prompt Engineering Toolbox

As part of this research, we developed the Autonomous Prompt Engineering Toolbox (APET). This toolbox is designed to enhance the effectiveness of prompts used with LLMs like GPT-4, thereby improving their performance and the quality of their outputs. The toolbox consists of a collection of advanced prompt engineering techniques that enable the LLM to select and apply the most appropriate methods based on the specific needs of a query.

The Autonomous Prompt Engineering Toolbox allows the LLM to operate more effectively across various tasks by leveraging refined inputs that reduce ambiguity, guide the model's focus, and clarify the intent of the query. This leads to responses that are not only more accurate but also contextually relevant, thereby enhancing the utility and adaptability of the LLM for a broad range of applications.

By incorporating different techniques such as expert prompting, chain of thought, and tree of thoughts, the toolbox enables the LLM to approach each query with a tailored strategy. These techniques provide structured guidance to the model, helping it navigate complex problem spaces more effectively and produce outputs that are not just correct, but also deep and insightful. Using the Autonomous Prompt Engineering Toolbox, LLMs can achieve a level of performance that approximates or even surpasses that of fine-tuned models for specific tasks without the need for extensive retraining.

---

[3] The term "Optimization" in this context refers to the process of refining the prompts to better align with the LLM's capabilities to interpret and respond. This adjustment process helps in maximizing the efficiency and accuracy of the model's responses, bridging the gap between a generic prompt and a tailored, more contextually aligned input that can evoke the best possible answer from the model.

### 3.3.2 Prompt Engineering techniques

The optimization process integrates several prompting techniques, each selected for their effectiveness in improving LLM outputs. The LLM is provided with a toolbox of prompt optimization techniques from which it can pick and choose what is more applicable for a certain prompt. These techniques, mostly influenced by the work of Wei et al. (2022), Xu et al. (2023), Yao et al. (2023), are central to the optimization strategy:

**Expert Prompting:** Adopts the model's ability to simulate expertise in specific domains, enhancing prompt quality and depth as suggested by Xu et al. (2023). This approach encourages the LLM to assume the role of an expert in the relevant domain, thereby tailoring its responses to reflect a level of understanding and insight that one would expect from a seasoned professional. This strategic personification is achieved by either explicitly instructing the model to adopt an expert persona. Such a technique leverages the LLM's inherent capacity for role-playing and contextual adaptation, drawing from its extensive pre-training on diverse genres and formats to simulate expert discourse.

Support for the efficacy of embedding an expert's persona within prompts is found in the foundational work of Brown et al. (2020) on GPT-3, which underscored the model's proficiency in few-shot learning, illustrating its capability to produce knowledgeable responses based on minimal examples. This highlights how expert prompting effectively focuses the model's attention, leveraging its few-shot learning capabilities to elicit more precise and authoritative outputs. Furthermore, Xu et al. (2023) have expanded upon this concept with their research into "Expert Prompting," which demonstrated that by constructing a prompt that positions the model as a domain-specific expert, the responses not only gain in accuracy but also reflect a depth and confidence akin to that of a human expert.

This approach leverages the principle of in-context learning, where models adjust their outputs based on the contextual cues provided by the expert identities. By simulating the depth and specificity of human experts, Expert Prompting enables LLMs to produce answers that are significantly lengthier and of higher quality compared to standard prompting methods. The methodological incorporation of expert identities ensures that the model's responses are tailored to the nuances of each query, reflecting a deeper understanding and specialization in the relevant domain.

**Chain of Thought (CoT)** works by structuring the response generation process into a series of logical, sequential steps. This method instructs the model to articulate its reasoning explicitly, mirroring the way humans approach problem-solving tasks. By breaking down the response into a coherent sequence of steps, CoT prompting enables the LLM to navigate complex queries with a structured and analytical approach, enhancing both the clarity and depth of the generated responses.

The effectiveness of the Chain of Thought approach is rooted in its ability to mimic human cognitive processes, guiding the LLM through a stepwise articulation of reasoning that leads to a more detailed and logically sound output. This technique is particularly effective for tasks that require complex reasoning or problem-solving capabilities. It prompts the model to engage in a more deliberate and systematic exploration of the query, leveraging its vast knowledge base in a more focused and organized manner. Wei et al. (2022) have underscored the significance of this approach, demonstrating how CoT prompting significantly improves LLM performance across a variety of reasoning tasks. By compelling the model to unpack the problem into manageable components and articulate each step of the thought process, CoT enhances the model's ability to generate solutions that are not only accurate but also explainable and aligned with logical reasoning patterns.

By adopting the Chain of Thought prompting, LLMs are encouraged to not only find solutions but also to provide a transparent reasoning trail that explains how they arrived at those solutions. This transparency in the reasoning process not only improves the interpretability of the model's outputs but also enhances trust in the model's capabilities to handle complex queries effectively. As a result, CoT prompting not only optimizes the accuracy and relevance of LLM responses but also contributes to the development of more sophisticated AI systems capable of engaging in nuanced and complex reasoning, mirroring the analytical depth characteristic of human thought processes.

**Tree of Thoughts (ToT)** enriches the reasoning capabilities of models by incorporating the dynamics of collaborative discussion among multiple expert personas. This sophisticated approach, originating from the research by Yao et al. (2023), builds upon and extends the "Chain of Thought" (CoT) methodology by introducing a multi-perspective dialogue that allows for an iterative and self-correcting reasoning process.

In ToT, each persona articulates a step of their reasoning and shares it with the group, creating a structured but fluid dialogue that mimics real-world expert deliberations. If any persona identifies a flaw in their reasoning, they withdraw their contribution, fostering a self-correcting mechanism that ensures only the most robust ideas prevail. This technique builds on the foundation laid by CoT, where the model is encouraged to detail its reasoning in a step-by-step manner. While CoT focuses on enhancing logical clarity and depth by unpacking the problem into manageable components, ToT takes this further by integrating multiple perspectives, thereby enriching the decision-making process with a broader range of insights and expertise.

Yao et al. (2023) conceptualized ToT to leverage the collective intelligence phenomenon, where diverse inputs from multiple 'experts' within the model can lead to more comprehensive and accurate problem-solving outputs. This approach significantly extends the single-threaded CoT by incorporating multiple threads of reasoning that interact and refine each other. Such a setup not only broadens the model's analytical perspective but also deepens its engagement with the problem, as it must consider and integrate diverse viewpoints and solutions.

In the autonomous optimization system envisioned for LLMs, the model provides the LLM with the theory around the selected prompting techniques, as depicted in Figure 2, to effectively enable the LLM to choose the optimal combination of prompting techniques for each specific query. This system equips the LLM with the flexibility to assess and decide which techniques from the toolbox—comprising Expert Prompting [1], Chain of Thought (CoT) [2], and Tree of Thoughts (ToT) [3]—are most suitable to enhance the clarity, depth, and relevance of its responses to individual prompts.

As each new sample prompt is inserted into the system, the LLM evaluates the nature of the query and its contextual requirements. It then autonomously selects from the techniques available, perhaps combining Expert Prompting to leverage domain-specific depth when the query demands expert-level discourse, or Chain of Thought to structure a clear, logical response pathway for complex problem-solving scenarios. For queries that benefit from diverse perspectives and collaborative refinement, the model might integrate the Tree of Thoughts approach, allowing for a multi-threaded analysis that enhances the robustness of the solution through iterative expert validation and correction.

By dynamically adapting its response strategy to the specifics of the prompt, the LLM can exploit the full potential of its vast training data, applying the most appropriate techniques to generate outputs that are tailored to meet the nuanced demands of each query. This adaptive capability pushes the boundaries of AI's problem-solving and reasoning capacities to closely mimic the sophisticated analytical processes typical of collective human expertise.

## System prompt

```
Imagine yourself as an  expert in the realm of prompting techniques for LLMs.
Your expertise is not just broad, encompassing the entire spectrum of current
knowledge on the subject, but also deep, delving into the nuances and
intricacies that many overlook. Your job is to reformulate prompts with surgical
precision, optimizing them for the most accurate response possible. The
reformulated prompt should enable the LLM to always give the correct answer to
the question.
```

## User prompt

```
Your available prompting techniques include, but are not limited to the
following:

- Crafting an expert who is an expert at the given task, by writing a high-
quality description about the most capable and suitable agent to answer the
instruction in second person perspective.[1]
- Explaining step-by-step how the problem should be tackled, and making sure
the model explains step-by-step how it came to the answer. You can do this by
adding "Let's think step-by-step".[2]
- Imagining three different experts who are discussing the problem at hand. All
experts will write down 1 step of their thinking, then share it with the group.
Then all experts will go on to the next step, etc. If any expert realises
they're wrong at any point then they leave.[3]
- Making sure all information needed is in the prompt, adding where necessary
but making sure the question remains having the same objective.

Your approach is methodical and analytical, yet creative. You use a mixture of
the prompting techniques, making sure you pick the right combination for each
instruction. You see beyond the surface of a prompt, identifying the core
objectives and the best ways to articulate them to achieve the desired outcomes.

Output instructions:""""
You should ONLY return the reformulated prompt. Make sure to include ALL
information from the given prompt to reformulate.
""""

Given above information and instructions, reformulate below prompt using the
techniques provided: """"
{sample_prompt}
""""
```

*Figure 2: Template of the instructions to the LLM used for producing Optimized Prompts, divided in the system prompt and user prompt.*

### 3.3.3 The Autonomous Prompt Engineering Toolbox: A Practical Example

This section illustrates the practical implementation of our prompt optimization techniques, showcasing their effectiveness in enhancing the model's output through a structured and expert-guided approach. We demonstrate this by analyzing a task from the Geometric Shapes benchmark dataset where the model must identify a geometric shape from an SVG path description. The correct answer to this task is "(G) pentagon".

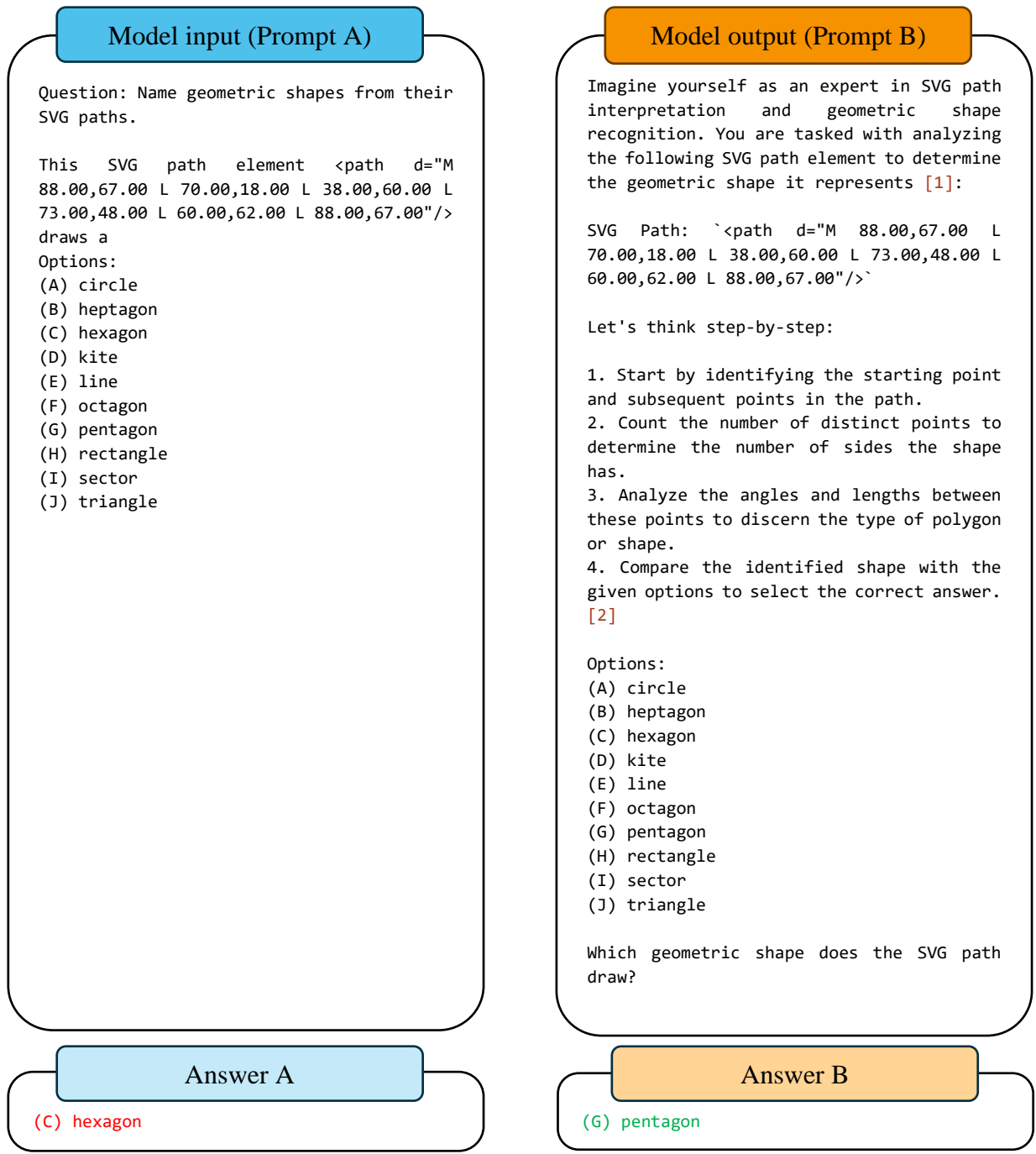

Figure 3: Practical example of the Prompt Optimization Toolbox in action.

The basic prompt provided to the model, identified by "Prompt A" in figure 3, is straightforward and relies heavily on the model's inherent reasoning capabilities. In this scenario, the lack of specific guidance may lead the model to misinterpret the task, resulting in the incorrect identification of the geometric shape. The response generated from this initial prompt incorrectly identified the shape as a "hexagon," which suggests that the model's internal reasoning was not aligned with the logical geometric analysis required.

To address the deficiencies of the initial prompt, the prompt engineering toolbox is applied, incorporating the "Expert Prompting", as seen at [1] and the Chain of Thought (CoT) method, as seen at [2]. The optimized prompt, identified by "Prompt B" in figure 3. systematically guides the model through the task and crafts a role that the model can take. The detailed step-by-step in this optimized prompt is designed to more accurately guide the model's processing, aligning it with the precise characteristics of geometric shapes. This structured approach mitigates the potential for misinterpretation seen in the original prompt and increases the likelihood of correctly identifying the geometric shape based on a logical analysis of the SVG path data, which it does correctly in the example.

### 3.4 Verification process

Upon completing the process, we compile a comprehensive dataset that captures a variety of data points. This dataset includes the following components:

- **Sample Prompt**: The initial question or statement provided to the model, serving as the baseline for response generation.
- **Optimized Prompt:** The refined version of the sample prompt, tailored through the prompt optimization techniques to potentially enhance the quality and specificity of the model's responses.
- **Benchmark Answer:** The correct response as defined by the benchmark dataset, used for evaluating the model's accuracy.
- **Answer Original:** The response generated from the original sample prompt, reflecting the model's capabilities without optimization.
- **Answer Optimized:** The response generated from the optimized prompt, intended to demonstrate the effects of prompt refinement.
- **Original Messages:** An array with the conversation with the LLM for the original prompt, showcasing how the interaction developed.
- **Optimized Messages:** The refined version of the messages array, containing the optimized prompt and the answers of the LLM to it.

The analysis of this dataset will involve statistical assessments of accuracy, comparing the model's chosen answers against the benchmark answers to quantify improvements or declines in performance attributable to the optimization process.

**3.5    Model Parameters**

In the experiment, the model was configured with a temperature setting of 0 to minimize randomness and enhance reproducibility. This setting is crucial because, although a temperature of 0 greatly reduces the probability of variable outputs, it does not entirely eliminate the chance that the model may generate different responses upon reprompting. This potential variability, even under controlled temperature settings, highlights the complexity of the model's operational dynamics.

To ensure thorough documentation and facilitate reproducibility, all experiment-related materials—including code, datasets, and detailed interaction logs—are openly shared on a dedicated [GitHub repository](#). This transparency allows other researchers to replicate the study precisely or to adapt the methodology for further exploration. It also supports the scientific community's broader validation efforts and fosters ongoing research advancements.

Additionally, it's important to note that other model parameters were left at their default settings during the experiments. These include 'top_p', which defaults to 1, allowing for a broader selection of tokens by considering the entire probability mass, and 'max_tokens', which limits the length of the model's outputs. These settings, along with temperature, are part of the model's configuration that influences its output characteristics—where adjusting either 'temperature' or 'top_p' (but not both) is recommended for controlling output variability. By adhering to these standard settings, the experiment maintains a balance between controlled reproducibility and the realistic application of the model, ensuring that the findings are both robust and applicable to real-world scenarios.

**4    Results**

This section covers the findings from an array of experiments designed to assess the autonomous optimization capabilities of GPT-4 within the framework of an optimization model, also referred to as the Autonomous Prompt Engineering Toolbox. This Toolbox provides the LLM with a suite of prompting techniques from which it autonomously learns and selects to enhance the standard prompts during the optimization step. The experiments tested the LLM's ability to improve its response quality across various tasks such as Word Sorting, Game Of 24, Geometric Shapes, and Checkmate in One by using these optimization tools.

The methodology, as detailed in the previous chapters, utilized the OpenAI API to implement this experiment in a controlled testing environment. Each task was first presented with a standard prompt and then with an optimized version generated by GPT-4 using the toolbox. This comparative approach aimed to quantitatively measure the performance enhancements attributable to the optimization process. The primary objective of this section is to present a comprehensive analysis of how the application of the toolbox influences overall performance across different tasks. The findings aim to underscore the effectiveness of autonomous prompt optimization in improving the accuracy and efficiency of GPT-4's responses. This examination not only highlights the capabilities of current AI systems in enhancing their operational efficacy autonomously but also sets the stage for a deeper exploration of individual prompting strategies within the toolbox and their specific impacts on model performance.

## 4.1 Overall Performance Enhancement

Table 1 presents a summary of the performance differences between standard and optimized prompts across the different tasks tested. The data reveal a general improvement in task performance when utilizing the Autonomous Prompt Engineering Toolbox (APET), with notable exceptions that warrant further investigation.

| Task | N | Standard | APET | Delta |
| --- | --- | --- | --- | --- |
| **Word Sorting** | 250 | 83.60% | 88.00% | +4.40 |
| **Game Of 24** | 75 | 16.00% | 18.67% | +2.67 |
| **Geometric Shapes** | 250 | 70.40% | 77.20% | +6.80 |
| **Checkmate in One** | 250 | 40.40% | 25.60% | -14.80 |

*Table 1: Performance Comparison Between Standard and Optimized Prompts*

In the task of Word Sorting, we observed a notable performance increase of 4.40%. This improvement can be attributed to the optimized prompts that are more aligned with GPT-4's natural language processing capabilities. Such refinements significantly enhance the model's ability to handle linguistic tasks by better structuring the information it processes.

The Game Of 24, which involves numerical reasoning, showed a moderate improvement of 2.67%. This increase suggests that the optimized prompts likely helped in more effectively structuring the problem-solving process for the LLM. By clarifying the task's requirements, these prompts enable GPT-4 to apply its numerical reasoning skills more efficiently.

For Geometric Shapes, a task that involves reasoning and simple geometrical calculations, there was a substantial increase of 6.80%—the highest improvement recorded among the tasks. This enhancement indicates that the optimization tools are particularly effective in framing the questions in a way that capitalizes on GPT-4's reasoning capabilities. By providing clearer task requirements, the prompts help the model navigate the complexities of spatial relationships more effectively.

Conversely, the task of Checkmate in One demonstrated a significant challenge, with a performance decline of 14.80%. This task, which requires strategic and spatial reasoning within the context of chess, may not be well-suited to the types of prompt optimizations that were applied. The decline suggests that the optimization techniques, while beneficial for more straightforward linguistic or numerical tasks, might oversimplify or fail to align with the complex strategic thinking required in chess. This misalignment highlights the need for a more nuanced approach to optimizing prompts for tasks that demand a high level of strategic depth and decision-making.

The observed improvements across most tasks suggest that GPT-4's performance is markedly enhanced by prompts that are better structured or contextualized to tap into its pre-existing knowledge and reasoning frameworks. This is consistent with findings in AI research indicating that even small changes in prompt formulation can significantly influence the performance of LLMs on specific tasks.

The decline in performance in the "Checkmate in One" task highlights a significant limitation in current optimization strategies. According to Kuo et al. (2023), natural language reasoning (NL reasoning) can help large language models (LLMs) generate a sense of "intent" but does not necessarily improve their performance. NL reasoning often introduces excessive and incorrect information into the model's analysis, leading to misguided strategies.

In the "Checkmate in One" task, allowing the model to use NL reasoning via CoT prompting likely led to the introduction of incorrect information, impairing its ability to make accurate strategic decisions. This issue is exacerbated by the model's tendency to hallucinate or fabricate details when dealing with complex scenarios. As a result, optimization techniques incorporating NL reasoning were detrimental to the model's performance in tasks requiring precise and accurate strategic thinking.

From a practical standpoint, these results reinforce the potential of using automated systems to enhance the effectiveness of LLMs across a variety of applications. By implementing prompt optimization, tasks that traditionally required significant human input to achieve high levels of accuracy can now see improved performance with less manual intervention. However, the results also caution against a one-size-fits-all approach to optimization, highlighting the importance of tailoring techniques to the specific cognitive requirements of each task.

### 4.2    Analysis of Prompting Techniques Usage

In this section, we present the usage statistics and effectiveness of various prompting techniques employed by GPT-4, facilitated through the Autonomous Prompt Engineering Toolbox. The data provided focuses on how frequently each technique is utilized (Table 2) and its impact on task performance (Table 3), forming a basis for the comprehensive analysis found in the discussion chapter.

| Task | Expert Only | CoT Only | ToT Only | Expert + CoT | Expert + ToT | CoT + ToT |
|---|---|---|---|---|---|---|
| **Word Sorting** | 18.00% | 0.40% | 0.00% | 81.60% | 0.00% | 0.00% |
| **Game Of 24** | 13.33% | 0.00% | 0.00% | 86.67% | 0.00% | 0.00% |
| **Geometric Shapes** | 0.40% | 59.20% | 1.20% | 38.80% | 0.00% | 0.40% |
| **Checkmate in One** | 4.40% | 16.80% | 1.60% | 72.80% | 0.00% | 4.00% |

*Table 2: Frequency of Usage of Prompting Techniques Across Tasks*

| Task | Expert Only | CoT Only | ToT Only | Expert + CoT | Expert + ToT | CoT + ToT |
|---|---|---|---|---|---|---|
| **Word Sorting** | 16.00% | 0.40% | 0.00% | 71.60% | 0.00% | 0.00% |
| **Game Of 24** | 1.33% | 0.00% | 0.00% | 17.33% | 0.00% | 0.00% |
| **Geometric Shapes** | 0.40% | 43.60% | 0.80% | 32.40% | 0.00% | 0.40% |
| **Checkmate in One** | 0.80% | 2.00% | 0.40% | 20.80% | 0.00% | 1.20% |

*Table 3: Effectiveness (% Correct of Total Correct) of Prompting Techniques Across Tasks*

The use of prompting techniques varies significantly across different tasks, reflecting GPT-4's ability to adapt its strategies to the unique requirements of each task. For Word Sorting, the model predominantly utilizes a combination of Expert Prompting and Chain of Thought (CoT), with this approach being applied in 81.60% of cases. This combination has proven to be highly effective, achieving a correctness rate of 71.60%, which is the highest among the techniques used for this task. The reliance on Expert Prompting and CoT suggests that the structured logical progression coupled with domain-specific knowledge significantly enhances the model's ability to process and organize linguistic information effectively.

In the Game of 24, there is a similar reliance on the combination of Expert Prompting and CoT, utilized in 86.67% of instances. However, the effectiveness for this task stands at 17.33%, indicating challenges in structured numerical problem-solving despite the logical sequencing provided by these techniques. This suggests that while the approach is favored, the complexity of the task poses substantial challenges, reflecting the need for perhaps more nuanced or advanced prompting strategies that could better handle the arithmetic complexities involved.

For tasks requiring the interpretation and visualization of spatial information, such as Geometric Shapes, CoT is heavily favored, either alone or in combination with Expert Prompting. CoT alone is used in 59.20% of cases, with its combination with Expert Prompting accounting for another 38.80%. The effectiveness of CoT alone stands at 43.60%, indicating that the step-by-step breakdown of tasks is particularly beneficial for managing spatial reasoning and visual processing.

The strategy shifts slightly for Checkmate in One, a task that combines the need for strategic foresight with a deep understanding of chess rules. Here, the combination of Expert Prompting and CoT is used in 72.80% of cases, achieving an effectiveness rate of 20.80%. This highlights how the integration of deep domain-specific knowledge with structured decision-making is crucial for tasks that involve complex strategic considerations.

Overall, the model's strategic use of prompting techniques showcases its capacity to align its computational strategies with the cognitive demands of various tasks. This adaptability is indicative of an advanced level of understanding, where GPT-4 not only selects but also effectively combines different techniques to optimize performance. The data presented here lays the groundwork for a deeper discussion on the effectiveness of these strategies, illustrating the model's capability to enhance task performance through sophisticated, dynamic prompting strategies.

## 5     Discussion

This chapter seeks to bridge the gap between the empirical findings and theoretical insights presented in earlier sections of this thesis. We will explore how our results compare with other existing methods and discuss the applicability of various prompting techniques to different tasks.

### 5.1     Evaluating Autonomous Prompt Engineering

The experimental results reveal significant insights into the performance of GPT-4 when utilizing self-optimized prompts compared to unoptimized ones. These findings are crucial for validating H1, which suggested that GPT-4 improves output quality significantly with self-optimized prompts. The data from tasks such as Word Sorting, Game of 24, and Geometric Shapes consistently demonstrated that the use of the Autonomous Prompt Engineering Toolbox (APET) leads to a marked improvement in performance. This enhancement can be attributed to the toolbox's ability to leverage GPT-4's inherent capabilities more effectively than standard prompting, thereby optimizing response quality across various tasks.

The autonomous prompt engineering toolbox demonstrates notable improvements in performance, paralleling advancements observed in leading-edge studies without relying on external data or specific training examples. Notably, Zhou et al. (2023) introduced the Automatic Prompt Engineer (APE), a system that enhances prompt efficiency by employing a heuristic search across a spectrum of generated instructions, markedly improving task performance. Unlike APE, which incorporates external inputs to refine prompts, our method employs an internally developed meta-prompting technique that reformulates initial prompts, facilitating enhanced clarity and focus without external data dependencies. This enables a more streamlined operation across various tasks, leading to responses that are not only precise but also contextually adept, thereby broadening the utility of LLMs for an extensive range of applications.

Similarly, Ye et al. (2024) reported the highest performance boost of 5.9% on the Big-Bench Hard benchmark through their example-driven prompt engineering methods. In contrast, our approach mainly equates these performance enhancements through zero-shot strategies, where the LLM internalizes and applies prompt engineering based on theoretical insights, negating the need for exemplar-based learning. This not only underscores our method's efficiency but also its ability to generalize across most tasks effectively.

Moreover, the methodology developed by Pryzant et al. (2023), which leverages textual gradients to refine prompts iteratively, highlights the potential for continuous prompt optimization. Our approach, while similarly adapting prompts, uniquely leverages the model's intrinsic capabilities to autonomously integrate and apply prompt engineering principles, achieving substantial performance gains.

Furthermore, Suzgun and Kalai (2024) explore meta-prompting, a technique that transcends traditional scaffolding by employing high-level, task-agnostic strategies to improve LLM functionality. This method significantly enhances LLM performance by managing and integrating multiple independent LM queries, similar in spirit to our use of meta-prompts, though their approach extends functionality through multiple iterations of the model and the integration of external tools like Python interpreters, which our system does not utilize.

Contrary to our expectations outlined in H3, the benefits of self-produced prompt optimization are not consistent across all prompt types, particularly in complex strategic tasks such as Checkmate in One. This task highlighted limitations in our prompting strategy, where it notably underperformed compared to three out of four other datasets. Unlike other methods, which maintain a mostly consistent performance across all tasks, our approach's deviation in this task underscores the need for further refinement to meet the nuanced demands of tasks requiring highly specialized knowledge or precise logical reasoning.

This inconsistency not only challenges the uniform applicability of self-optimization, as posited in H3, but also brings us to reconsider H2. While self-optimization can match or even surpass external methods in some contexts, the evidence for H2 is only partial. Our findings reveal that although internal optimizations are highly effective, their performance equivalence with specialized external methods varies significantly by task. This variation suggests a crucial need for a more adaptive and scalable approach in the self-optimization techniques employed by GPT-4. Collectively, these comparisons do demonstrate that our method not only largely aligns with the state-of-the-art advancements but also introduces an innovative approach in prompt engineering by reducing dependencies on external enhancements. This breakthrough not only amplifies the practical applicability of LLMs across a diverse spectrum of tasks but also deepens our comprehension of how such models can autonomously leverage

## 5.2 Alignment of Prompting Techniques with Task Requirements

This section explores the strategic deployment of prompting techniques—Expert Prompting, Chain of Thought (CoT), and Tree of Thoughts (ToT)—across diverse cognitive tasks such as Checkmate in One Move, Geometric Shapes, Game of 24, and Word Sorting[4]. We will also discuss the notable minimal use of ToT and why Geometric Shapes differs markedly in its use of prompting techniques compared to other tasks.

In the Checkmate in One Move task, the predominant use of Expert Prompting combined with CoT (72.80% usage) aligns with the task's requirements for deep knowledge of chess rules and strategic foresight. Expert Prompting taps into the model's potential training on chess data, enhancing its ability to anticipate plausible moves. CoT complements this by methodically guiding the model through the decision-making process, crucial for strategizing successful checkmates (BIG-Bench authors, 2023; Wei et al., 2022; Xu et al., 2023).

Geometric Shapes predominantly utilizes CoT (59.20% usage), reflecting its demand for sequential logical processing to interpret SVG paths and construct visual representations. This task's reliance on procedural knowledge rather than deep, domain-specific expertise explains the relative minimal use of Expert Prompting and almost no use of ToT (0.40%). The unique characteristics of this task—requiring the translation of textual commands into visual forms—makes it distinct from others, where domain knowledge or more complex decision trees might be more pertinent (Suzgun et al., 2023; Wei et al., 2022).

The Game of 24, which leverages CoT supported by Expert Prompting (86.67% combined usage), highlights the task's emphasis on mathematical and logical reasoning. CoT aids in decomposing the problem into manageable arithmetic steps, while Expert Prompting enhances the model's efficiency and strategic approach in reaching the target number (Yao et al., 2023; Wei et al., 2022; Xu et al., 2023).

Word Sorting heavily employs a combination of Expert Prompting and CoT (81.60%), integrating linguistic comprehension with algorithmic execution. This approach is effective in sorting tasks, reflecting the dual necessity for understanding linguistic rules and applying them in a structured, logical sequence (Suzgun et al., 2023; Wei et al., 2022; Xu et al., 2023).

Across these tasks, ToT is notably underutilized, indicating that the scenarios presented do not frequently require the exploration of multiple complex reasoning pathways simultaneously, which ToT is designed to facilitate. The minimal application suggests that the tasks predominantly benefit from either deep expertise or structured sequential processing, rather than the need for navigating through multiple potential solutions concurrently (Yao et al., 2023).

---

[4] **Note**: Given the black-box nature of LLMs, it is difficult to definitively ascertain why a model favours certain prompting techniques over others for specific tasks. However, through an examination of the cognitive requirements of the tasks and the theoretical foundations of various prompting techniques, we can hypothesize plausible explanations for these preferences, offering insights into the alignment between task demands and the capabilities these techniques enhance.

This analysis underscores how the choice of prompting techniques is linked to the specific cognitive demands of each task. By appropriately matching these techniques to tasks, the LLM achieves better performance, showcasing the effective application of cognitive theories and AI research in practical settings. The distinction in the use of prompting techniques, particularly the unique case of Geometric Shapes and the minimal deployment of ToT, highlights the importance of task-specific strategy formulation in the deployment of LLM capabilities.

# 6   Conclusion

## 6.1   Overall Summary

This paper investigated the capabilities of GPT-4 with a focus on its ability to autonomously enhance its performance through self-optimized prompting, employing the Autonomous Prompt Engineering Toolbox (APET). The overarching goal was to determine the extent to which the LLM could apply theoretical insights autonomously to improve its performance across a spectrum of tasks. This research has not only provided a deep dive into the internal mechanics of prompt optimization but also filled a significant gap in understanding how a large language model can dynamically apply learned strategies to optimize its outputs without external intervention.

The study yielded several critical findings regarding the self-optimization capabilities of GPT-4, leading to valuable academic and scientific insights. First and foremost, the paper confirmed that self-optimized prompts substantially improve task performance. Specifically, tasks that involved linguistic processing and simple logical reasoning like Word Sorting and Geometric Shapes showed significant performance enhancements when utilizing self-optimized prompts. This indicates that GPT-4 can effectively apply theoretical optimization strategies to real-world tasks, enhancing its performance autonomously.

Second, while GPT-4 demonstrated a notable ability to optimize its responses effectively across many tasks, the benefits of such optimizations were not uniformly realized across all types of tasks. In more complex strategic tasks, such as Checkmate in One, the model struggled to apply optimization strategies effectively, highlighting the challenges of applying a one-size-fits-all optimization strategy across diverse cognitive domains.

Third, this study shed light on the alignment of specific prompting techniques with task requirements, demonstrating how the strategic deployment of techniques such as Expert Prompting and Chain of Thought can be crucial in enhancing the model's performance. The findings indicate that GPT-4 can not only adopt but also effectively adapt these strategies to the cognitive demands of different tasks, showcasing its flexibility and the potential for autonomous task-specific strategy formulation.

Thus, the observed decline in performance for the "Checkmate in one" task suggests that H4 is only partially validated. This outcome indicates that while GPT-4 demonstrates a substantial capacity for autonomous operation, it simultaneously highlights specific areas necessitating further refinement. However, the research does underscore the substantial potential for large language models to advance towards greater autonomy in their operational capabilities. By successfully applying internalized strategies for prompt optimization, the APET has shown that it can extend the boundaries of what is possible with autonomous AI systems in real-world applications.

In conclusion, the paper provides compelling evidence that GPT-4's self-optimization capabilities enable it to apply theoretical insights from the Toolbox to improve its performance across a variety of tasks. This capability marks a significant step forward in the development of autonomous AI systems, suggesting that future AI models could be designed to not only perform tasks but also self-optimize in real-time to adapt to new challenges dynamically.

## 6.2  Limitations and Future Research

The findings of this thesis, while illuminating the capabilities of GPT-4 in terms of autonomous prompt optimization, are circumscribed by several inherent limitations which naturally point towards areas for further research. Notably, the study's exploration was confined to a limited set of cognitive tasks, which, while diverse, do not encompass the full spectrum of challenges that large language models might encounter in real-world applications. This limitation underscores the necessity for extending the variety of tasks in future studies to include more complex and varied scenarios, providing a more comprehensive assessment of the model's capabilities.

Another significant constraint was the range of prompting techniques tested. The study primarily focused on a select few combinations of prompting strategies, potentially overlooking other effective combinations or sequences that might yield different insights into the model's operational dynamics. Future research should, therefore, consider a broader array of prompting techniques and their permutations to fully explore the space of possible optimizations and their impacts on model performance.

The research methodology itself, while robust, is tailored specifically to the version of GPT-4 used and the specific tasks at hand. As such, the generalizability of the findings may be limited. To mitigate this, subsequent studies could look to replicate the experiments across different versions of GPT or other large language models. Such cross-model testing could help validate the findings and establish a more generalized understanding of prompt optimization capabilities across the AI field.

Looking forward, there is a rich vein of potential research that could build on the groundwork laid by this thesis. Investigating how different models handle a wider array of prompt types and task complexities could further clarify the limits and potential of autonomous optimization. Moreover, integrating these strategies into real-world applications and studying their effectiveness and adaptability over time would provide invaluable insights into the practical utility and scalability of self-optimizing AI systems.

Furthermore, enhancing the theoretical models that underpin prompt optimization could lead to more sophisticated systems capable of more nuanced understanding and interaction with human users. This could involve deeper integration of cognitive and computational theories to refine the models' decision-making processes, potentially leading to breakthroughs in AI's ability to understand and respond to complex human needs and contexts.

In conclusion, while this research has taken significant strides in understanding and demonstrating the potential of GPT-4 for autonomous prompt optimization, the path forward is ripe with opportunities for deeper exploration and refinement. Expanding the scope of tasks, exploring a wider range of prompting techniques, and applying these insights in practical, real-world contexts are essential next steps for advancing the field and fully realizing the transformative potential of large language models.

# 7 References

Arora, S., Narayan, A., Chen, M. F., Orr, L., Guha, N., Bhatia, K., Chami, I., Sala, F., & Ré, C. (2022). *Ask Me Anything: A simple strategy for prompting language models* (arXiv:2210.02441). arXiv. https://doi.org/10.48550/arXiv.2210.02441

Bahdanau, D., Cho, K., & Bengio, Y. (2016). *Neural Machine Translation by Jointly Learning to Align and Translate* (arXiv:1409.0473). arXiv. https://doi.org/10.48550/arXiv.1409.0473

Bommasani, R., Hudson, D. A., Adeli, E., Altman, R., Arora, S., von Arx, S., Bernstein, M. S., Bohg, J., Bosselut, A., Brunskill, E., Brynjolfsson, E., Buch, S., Card, D., Castellon, R., Chatterji, N., Chen, A., Creel, K., Davis, J. Q., Demszky, D., … Liang, P. (2022). *On the Opportunities and Risks of Foundation Models* (arXiv:2108.07258). arXiv. https://doi.org/10.48550/arXiv.2108.07258

Brown, T. B., Mann, B., Ryder, N., Subbiah, M., Kaplan, J., Dhariwal, P., Neelakantan, A., Shyam, P., Sastry, G., Askell, A., Agarwal, S., Herbert-Voss, A., Krueger, G., Henighan, T., Child, R., Ramesh, A., Ziegler, D. M., Wu, J., Winter, C., … Amodei, D. (2020). *Language Models are Few-Shot Learners* (arXiv:2005.14165). arXiv. https://doi.org/10.48550/arXiv.2005.14165

Bubeck, S., Chandrasekaran, V., Eldan, R., Gehrke, J., Horvitz, E., Kamar, E., Lee, P., Lee, Y. T., Li, Y., Lundberg, S., Nori, H., Palangi, H., Ribeiro, M. T., & Zhang, Y. (2023). *Sparks of Artificial General Intelligence: Early experiments with GPT-4* (arXiv:2303.12712). arXiv. https://doi.org/10.48550/arXiv.2303.12712

Chen, B., Zhang, Z., Langrené, N., & Zhu, S. (2023). *Unleashing the potential of prompt engineering in Large Language Models: A comprehensive review* (arXiv:2310.14735). arXiv. http://arxiv.org/abs/2310.14735

Cheng, D., Huang, S., Bi, J., Zhan, Y., Liu, J., Wang, Y., Sun, H., Wei, F., Deng, D., & Zhang, Q. (2023). *UPRISE: Universal Prompt Retrieval for Improving Zero-Shot Evaluation* (arXiv:2303.08518). arXiv. https://doi.org/10.48550/arXiv.2303.08518

Collins, K. M., Wong, C., Feng, J., Wei, M., & Tenenbaum, J. B. (2022). *Structured, flexible, and robust: Benchmarking and improving large language models towards more human-like behavior in out-of-distribution reasoning tasks* (arXiv:2205.05718). arXiv. https://doi.org/10.48550/arXiv.2205.05718

Devlin, J., Chang, M.-W., Lee, K., & Toutanova, K. (2019). *BERT: Pre-training of Deep Bidirectional Transformers for Language Understanding* (arXiv:1810.04805). arXiv. https://doi.org/10.48550/arXiv.1810.04805

Du, Y., Li, S., Torralba, A., Tenenbaum, J. B., & Mordatch, I. (2023). *Improving Factuality and Reasoning in Language Models through Multiagent Debate* (arXiv:2305.14325). arXiv. https://doi.org/10.48550/arXiv.2305.14325

Hulbert, D. (2023). *Using Tree-of-Thought Prompting to boost ChatGPT's reasoning* (0.1) [Computer software]. https://github.com/dave1010/tree-of-thought-prompting (Original work published 2023)

Kojima, T., Gu, S. S., Reid, M., Matsuo, Y., & Iwasawa, Y. (2023). *Large Language Models are Zero-Shot Reasoners* (arXiv:2205.11916). arXiv. https://doi.org/10.48550/arXiv.2205.11916

Kuo, M.-T., Hsueh, C.-C., & Tsai, R. T.-H. (2023). *Large Language Models on the Chessboard: A Study on ChatGPT's Formal Language Comprehension and Complex Reasoning Skills* (arXiv:2308.15118). arXiv. http://arxiv.org/abs/2308.15118

Lee, T. B., & Trott, S. (2023, December 7). *Large language models, explained with a minimum of math and jargon*. https://www.understandingai.org/p/large-language-models-explained-with

Li, W., Li, L., Xiang, T., Liu, X., Deng, W., & Garcia, N. (2024). *Can multiple-choice questions really be useful in detecting the abilities of LLMs?* (arXiv:2403.17752). arXiv. http://arxiv.org/abs/2403.17752

Li, Y., Wei, F., Zhao, J., Zhang, C., & Zhang, H. (2023). *RAIN: Your Language Models Can Align Themselves without Finetuning* (arXiv:2309.07124). arXiv. http://arxiv.org/abs/2309.07124

Lin, S., Hilton, J., & Evans, O. (2022). *TruthfulQA: Measuring How Models Mimic Human Falsehoods* (arXiv:2109.07958). arXiv. https://doi.org/10.48550/arXiv.2109.07958

Milmo, D. (2023, February 2). ChatGPT reaches 100 million users two months after launch. *The Guardian*. https://www.theguardian.com/technology/2023/feb/02/chatgpt-100-million-users-open-ai-fastest-growing-app

Moskvichev, A., Odouard, V. V., & Mitchell, M. (2023). *The ConceptARC Benchmark: Evaluating Understanding and Generalization in the ARC Domain* (arXiv:2305.07141). arXiv. https://doi.org/10.48550/arXiv.2305.07141

Naveed, H., Khan, A. U., Qiu, S., Saqib, M., Anwar, S., Usman, M., Akhtar, N., Barnes, N., & Mian, A. (2023). *A Comprehensive Overview of Large Language Models* (arXiv:2307.06435). arXiv. http://arxiv.org/abs/2307.06435

Nori, H., Lee, Y. T., Zhang, S., Carignan, D., Edgar, R., Fusi, N., King, N., Larson, J., Li, Y., Liu, W., Luo, R., McKinney, S. M., Ness, R. O., Poon, H., Qin, T., Usuyama, N., White, C., & Horvitz, E. (2023). *Can Generalist Foundation Models Outcompete Special-Purpose Tuning? Case Study in Medicine* (arXiv:2311.16452). arXiv. https://doi.org/10.48550/arXiv.2311.16452

OpenAI, Achiam, J., Adler, S., Agarwal, S., Ahmad, L., Akkaya, I., Aleman, F. L., Almeida, D., Altenschmidt, J., Altman, S., Anadkat, S., Avila, R., Babuschkin, I., Balaji, S., Balcom, V., Baltescu, P., Bao, H., Bavarian, M., Belgum, J., … Zoph, B. (2023). *GPT-4 Technical Report* (arXiv:2303.08774). arXiv. http://arxiv.org/abs/2303.08774

Pang, J.-C., Wang, P., Li, K., Chen, X.-H., Xu, J., Zhang, Z., & Yu, Y. (2023). *Language Model Self-improvement by Reinforcement Learning Contemplation*.

Pezeshkpour, P., & Hruschka, E. (2023). *Large Language Models Sensitivity to The Order of Options in Multiple-Choice Questions* (arXiv:2308.11483). arXiv. https://doi.org/10.48550/arXiv.2308.11483

Pryzant, R., Iter, D., Li, J., Lee, Y., Zhu, C., & Zeng, M. (2023). Automatic Prompt Optimization with "Gradient Descent" and Beam Search. In H. Bouamor, J. Pino, & K. Bali (Eds.), *Proceedings of the 2023 Conference on Empirical Methods in Natural Language Processing* (pp. 7957–7968). Association for Computational Linguistics. https://doi.org/10.18653/v1/2023.emnlp-main.494

Radford, A., Narasimhan, K., Salimans, T., & Sutskever, I. (2018). *Improving Language Understanding by Generative Pre-Training*.

Ren, J., Zhao, Y., Vu, T., Liu, P. J., & Lakshminarayanan, B. (2023). *Self-Evaluation Improves Selective Generation in Large Language Models* (arXiv:2312.09300). arXiv. https://doi.org/10.48550/arXiv.2312.09300

Srivastava, A., Rastogi, A., Rao, A., Shoeb, A. A. M., Abid, A., Fisch, A., Brown, A. R., Santoro, A., Gupta, A., Garriga-Alonso, A., Kluska, A., Lewkowycz, A., Agarwal, A., Power, A., Ray, A., Warstadt, A., Kocurek, A. W., Safaya, A., Tazarv, A., … Wu, Z. (2023). *Beyond the Imitation Game: Quantifying and extrapolating the capabilities of language models* (arXiv:2206.04615). arXiv. https://doi.org/10.48550/arXiv.2206.04615

Sun, H., Li, X., Xu, Y., Homma, Y., Cao, Q., Wu, M., Jiao, J., & Charles, D. (2023). *AutoHint: Automatic Prompt Optimization with Hint Generation* (arXiv:2307.07415). arXiv. https://doi.org/10.48550/arXiv.2307.07415

Suzgun, M., & Kalai, A. T. (2024). *Meta-Prompting: Enhancing Language Models with Task-Agnostic Scaffolding* (arXiv:2401.12954). arXiv. http://arxiv.org/abs/2401.12954

Suzgun, M., Scales, N., Schärli, N., Gehrmann, S., Tay, Y., Chung, H. W., Chowdhery, A., Le, Q. V., Chi, E. H., Zhou, D., & Wei, J. (2022). *Challenging BIG-Bench Tasks and Whether Chain-of-Thought Can Solve Them* (arXiv:2210.09261). arXiv. https://doi.org/10.48550/arXiv.2210.09261

*Understanding searches better than ever before*. (2019, October 25). Google. https://blog.google/products/search/search-language-understanding-bert/

Vaswani, A., Shazeer, N., Parmar, N., Uszkoreit, J., Jones, L., Gomez, A. N., Kaiser, L., & Polosukhin, I. (2023). *Attention Is All You Need* (arXiv:1706.03762). arXiv. https://doi.org/10.48550/arXiv.1706.03762

Wang, A., Pruksachatkun, Y., Nangia, N., Singh, A., Michael, J., Hill, F., Levy, O., & Bowman, S. R. (2020). *SuperGLUE: A Stickier Benchmark for General-Purpose Language Understanding Systems* (arXiv:1905.00537). arXiv. https://doi.org/10.48550/arXiv.1905.00537

Wang, A., Singh, A., Michael, J., Hill, F., Levy, O., & Bowman, S. R. (2019). *GLUE: A Multi-Task Benchmark and Analysis Platform for Natural Language Understanding* (arXiv:1804.07461). arXiv. https://doi.org/10.48550/arXiv.1804.07461

Wang, X., Wei, J., Schuurmans, D., Le, Q., Chi, E., Narang, S., Chowdhery, A., & Zhou, D. (2023). *Self-Consistency Improves Chain of Thought Reasoning in Language Models* (arXiv:2203.11171). arXiv. https://doi.org/10.48550/arXiv.2203.11171

Wei, J., Tay, Y., Bommasani, R., Raffel, C., Zoph, B., Borgeaud, S., Yogatama, D., Bosma, M., Zhou, D., Metzler, D., Chi, E. H., Hashimoto, T., Vinyals, O., Liang, P., Dean, J., & Fedus, W. (2022). *Emergent Abilities of Large Language Models* (arXiv:2206.07682). arXiv. https://doi.org/10.48550/arXiv.2206.07682

Wei, J., Wang, X., Schuurmans, D., Bosma, M., Ichter, B., Xia, F., Chi, E., Le, Q., & Zhou, D. (2023). *Chain-of-Thought Prompting Elicits Reasoning in Large Language Models* (arXiv:2201.11903). arXiv. https://doi.org/10.48550/arXiv.2201.11903

Wu, T., Terry, M., & Cai, C. J. (2022). *AI Chains: Transparent and Controllable Human-AI Interaction by Chaining Large Language Model Prompts* (arXiv:2110.01691). arXiv. https://doi.org/10.48550/arXiv.2110.01691

Xu, B., Yang, A., Lin, J., Wang, Q., Zhou, C., Zhang, Y., & Mao, Z. (2023). *ExpertPrompting: Instructing Large Language Models to be Distinguished Experts* (arXiv:2305.14688). arXiv. http://arxiv.org/abs/2305.14688

Yao, S., Yu, D., Zhao, J., Shafran, I., Griffiths, T. L., Cao, Y., & Narasimhan, K. (2023). *Tree of Thoughts: Deliberate Problem Solving with Large Language Models* (arXiv:2305.10601). arXiv. https://doi.org/10.48550/arXiv.2305.10601

Ye, Q., Axmed, M., Pryzant, R., & Khani, F. (2024). *Prompt Engineering a Prompt Engineer* (arXiv:2311.05661). arXiv. https://doi.org/10.48550/arXiv.2311.05661

Zamfirescu-Pereira, J. D., Wong, R., Hartmann, B., & Yang, Q. (2023, April 23). *Why Johnny Can't Prompt: How Non-AI Experts Try (and Fail) to Design LLM Prompts*. https://doi.org/10.1145/3544548.3581388

Zhong, W., Cui, R., Guo, Y., Liang, Y., Lu, S., Wang, Y., Saied, A., Chen, W., & Duan, N. (2023). *AGIEval: A Human-Centric Benchmark for Evaluating Foundation Models* (arXiv:2304.06364). arXiv. https://doi.org/10.48550/arXiv.2304.06364

Zhou, Y., Muresanu, A. I., Han, Z., Paster, K., Pitis, S., Chan, H., & Ba, J. (2023). *Large Language Models Are Human-Level Prompt Engineers* (arXiv:2211.01910). arXiv. https://doi.org/10.48550/arXiv.2211.01910